\documentclass[conference]{IEEEtran}
\IEEEoverridecommandlockouts                   
\usepackage{times}
\usepackage{graphicx}
\usepackage{subcaption}
\usepackage[numbers]{natbib}
\usepackage{multicol}
\usepackage[bookmarks=true,hidelinks]{hyperref}
\usepackage{amsmath}
\usepackage{amssymb}
\usepackage{graphicx}
\newcommand{\norm}[1]{\lVert #1 \rVert}

\begin{document}

\title{Learning Physical Systems: Symplectification via Gauge Fixing in Dirac Structures}



%
\author{%
  Aristotelis Papatheodorou\textsuperscript{*},
  Pranav Vaidhyanathan\textsuperscript{*},
  Natalia Ares,
  Ioannis Havoutis%
  \\[1ex]
  Department of Engineering Science, University of Oxford, Oxford, OX1 3PJ, United Kingdom%
  \thanks{\textsuperscript{*}These authors contributed equally to this work.}%
  \thanks{Correspondence to: \texttt{\{aristotelis, pranav\}@robots.ox.ac.uk}}%
  \thanks{Paper presented at "Equivariant Systems: Theory and Applications in State Estimation, Artificial Intelligence and Control", Robotics: Science and Systems (RSS) 2025 Workshop.}
}

\maketitle

\begin{abstract}
Physics-informed deep learning has achieved remarkable progress by embedding geometric priors, such as Hamiltonian symmetries and variational principles, into neural networks, enabling structure-preserving models that extrapolate with high accuracy. However, in systems with dissipation and holonomic constraints, ubiquitous in legged locomotion and multibody robotics, the canonical symplectic form becomes degenerate, undermining the very invariants that guarantee stability and long-term prediction. In this work, we tackle this foundational limitation by introducing Presymplectification Networks (PSNs), the first framework to learn the symplectification lift via Dirac structures, restoring a non-degenerate symplectic geometry by embedding constrained systems into a higher-dimensional manifold. Our architecture combines a recurrent encoder with a flow-matching objective to learn the augmented phase-space dynamics end-to-end. We then attach a lightweight Symplectic Network (SympNet) to forecast constrained trajectories while preserving energy, momentum, and constraint satisfaction. We demonstrate our method on the dynamics of the ANYmal quadruped robot, a challenging contact-rich, multibody system. To the best of our knowledge, this is the first framework that effectively bridges the gap between constrained, dissipative mechanical systems and symplectic learning, unlocking a whole new class of geometric machine learning models, grounded in first principles yet adaptable from data.

\end{abstract}

\IEEEpeerreviewmaketitle

\section{Introduction}
Physics-Informed deep learning has seen remarkable success in recent years, with Physics Informed Neural Networks (PINNs)~\cite{PINNs} excelling in different scientific fields~\cite{KIYANI2025117984, karniadakis2021physics,li2020fourier,vaidhyanathan2025metasym}. This advancement is due to the development of neural architectures that incorporate geometric structures inherent in physical systems, such as Hamiltonian, Lagrangian, or port-Hamiltonian forms. These structures enable these architectures to conserve energy and momentum, extend their predictive capabilities far beyond the training data, and learn from relatively limited datasets \cite{goswami2022physicsinformeddeepneuraloperator}. Notable examples include Hamiltonian Neural Networks (HNNs) \cite{greydanus2019hamiltonianneuralnetworks}, Lagrangian Neural Networks (LNNs)~\cite{cranmer2020lagrangianneuralnetworks} as well as Symplectic Networks (SympNets)~\cite{jin2020sympnets}, together with more recent Symplectic Neural Flows~\cite{canizares2024symplecticneuralflowsmodeling} and port-Hamiltonian ODE nets~\cite{duong2024porthamiltonianneuralodenetworks}. At the heart of these approaches and virtually in almost all aspects of physics, lies the canonical symplectic two form,
\begin{equation*}
    \omega=d q^i \wedge d p_i,
\end{equation*}
defined on the cotangent bundle $T^* Q$ of a configuration manifold $Q$. Since $\omega$ is both \textit{closed} ($d\omega=0$) and non-degenerate (its bilinear pairing has no null directions), this supplies the geometric machinery that turns a Hamiltonian $H(\mathbf q, \mathbf p)$ into a vector field via $X_H \lrcorner \omega =d H$ \cite{arnold2013mathematical}. This construction guarantees phase-space volume preservation (Liouville’s theorem) and locks in first integrals such as total energy and linear/angular momentum that are vital for long-horizon prediction with sparse data.

\begin{figure*}[ht!]
  \centering
  \begin{subfigure}[t]{0.4\textwidth}
    \centering
    \includegraphics[width=\linewidth]{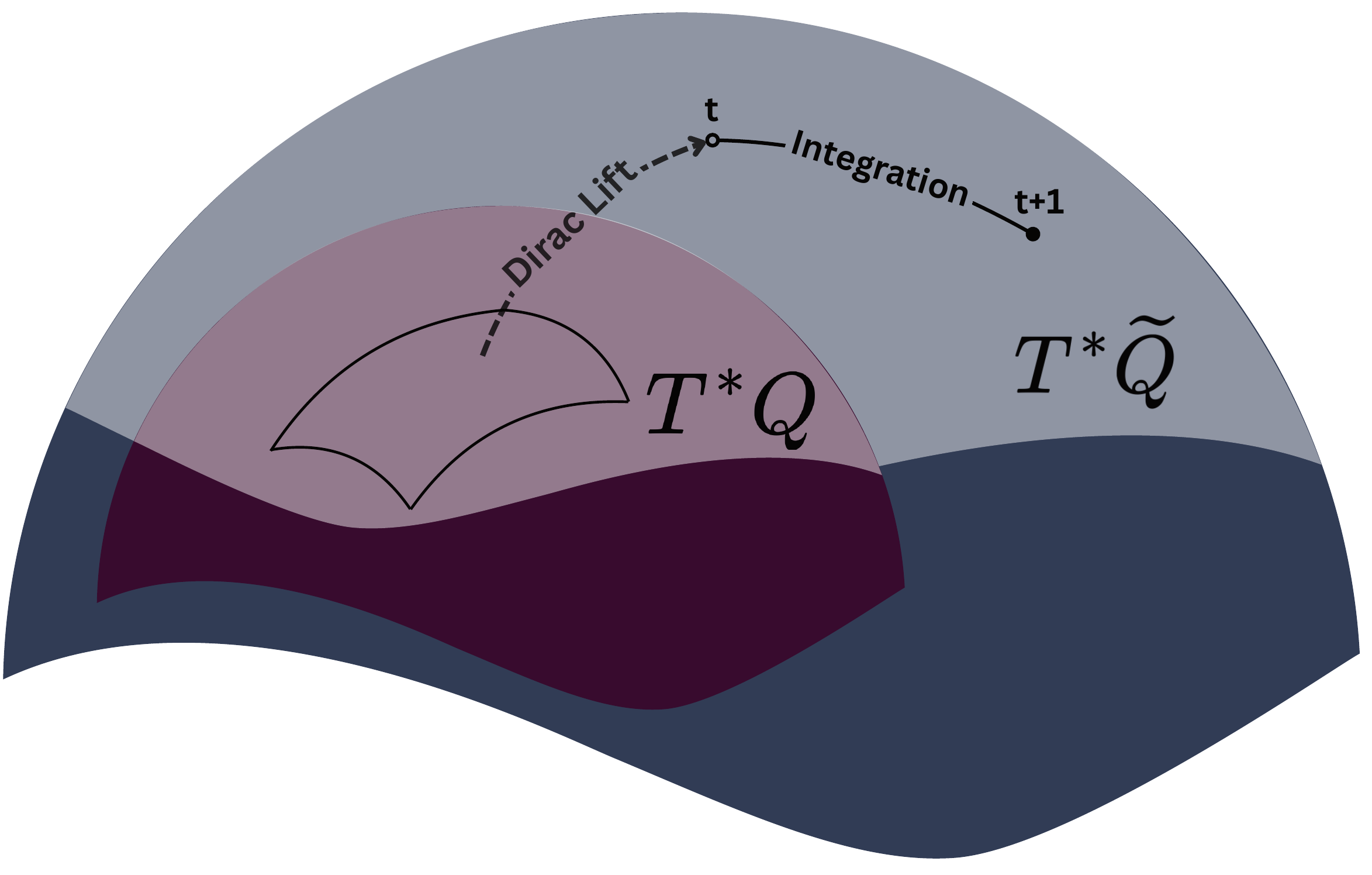}
    \caption{}
  \end{subfigure}
  \hfill
  \begin{subfigure}[t]{0.58\textwidth}
    \centering
    \includegraphics[width=\linewidth]{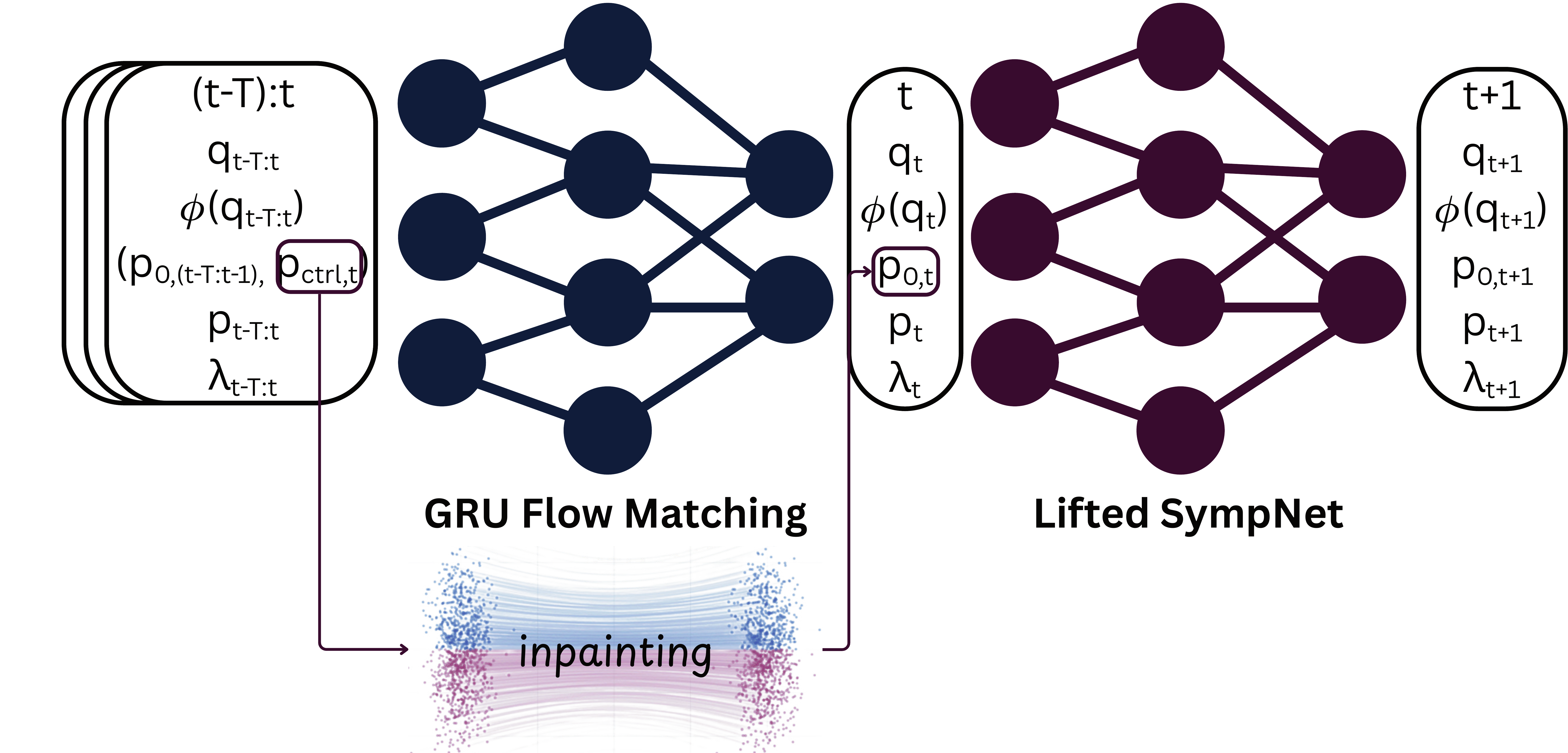}
    \caption{}
  \end{subfigure}
    \caption{(a) Our framework lifts the original phase-space to a fully conservative higher-dimensional manifold using the Dirac-lifted symplectification procedure. Then the dynamics are integrated along the surface of the lifted manifold. (b) The Dirac-lifted symplectification procedure is implemented as a flow-matching, inpainting objective using a GRU. The procedure maps the control-induced conjugate momenta ($\mathbf p_{ctrl,t}$) to the total conjugate momenta that include the system's dissipation ($\mathbf p_{0,t}$). An additional context ($T$) of 10 timesteps is supplied to the network. Then, the lifted phase-space is supplied to a SympNet that performs the next timestep prediction.}
  \label{fig:framework_overview}
\end{figure*}

However, many real-world mechanisms violate this assumption. Systems with rigid holonomic constraints and foot–ground contacts degenerate the symplectic form, splitting phase space into uncontrolled directions where invariants are no longer protected. In practice, physics informed models confronted with contacts or closed kinematic chains suffer energy blow-up, constraint drift, and brittle generalization. Soft-penalty methods can reduce, but never eliminate, these pathologies so far \cite{xiong2022nonseparablesymplecticneuralnetworks}. However, in their seminal work on symplectic optimization, França \textit{et al.} \cite{frança2023optimizationmanifoldssymplecticapproach}, introduced the concept of symplectification in the context of integrators. \textit{Symplectification} refers to the process of {recovering a non-degenerate symplectic form from a degenerate one, by embedding the original phase space into a higher-dimensional manifold}. Concretely, given a presymplectic manifold $(\mathcal{S},\omega)$ whose 2-form $\omega$ has rank deficiencies induced by holonomic constraints, we can attach:
\begin{itemize}
    \item  a clock coordinate $q^{0}=t$ with conjugate momentum $p_{0}$, and
    \item the Lagrange multipliers $\lambda_{a}$ together with their conjugate momenta $\pi_{a}$.
\end{itemize}  

The resulting extended bundle:
~
\begin{equation}
\begin{aligned}
&T^{*}\widetilde{Q} := T^{*}Q \times T^{*}\!\bigl(\mathbb{R}\times\mathbb{R}^{m}\bigr),\\
&(\mathbf{Q},\mathbf{P}) := (q^{0},q^{i},\lambda_{a};\,p_{0},p_{i},\pi_{a}),
\end{aligned}
\end{equation}
~
carries the {canonical symplectic form}:
~
\begin{equation}
\widetilde\Omega = dq^{0}\wedge dp_{0} + dq^{i}\wedge dp_{i} + d\lambda_{a}\wedge d\pi_{a},
\end{equation}
which is closed and \emph{non-degenerate} by construction. Choosing the {extended Hamiltonian} $\widetilde{H}(\mathbf Q, \mathbf P) = H(\mathbf q, \mathbf p) + p_{0} + \lambda_{a}\,\phi^{a}(\mathbf q)$, where $\frac{d\pi_a }{dt}= -\phi^a(\mathbf q)=0$ encodes the original constraints, Hamilton's equations on $(T^{*}\widetilde{Q},\widetilde\Omega)$ reproduce the constrained dynamics once the \emph{Dirac gauge}:
\begin{equation}
    q^{0}=t,\; p_{0}+H+\lambda_{a}\phi^{a}=0,\; \pi_{a}=0,
\end{equation}
is imposed. In effect, symplectification trades degeneracy for auxiliary coordinates, thereby {restoring the geometric backbone} needed for structure-preserving neural networks \cite{gotay1979presymplectic, libermann2012symplectic}.

\begin{figure*}[htb!]
    \includegraphics[width=1.0\textwidth]{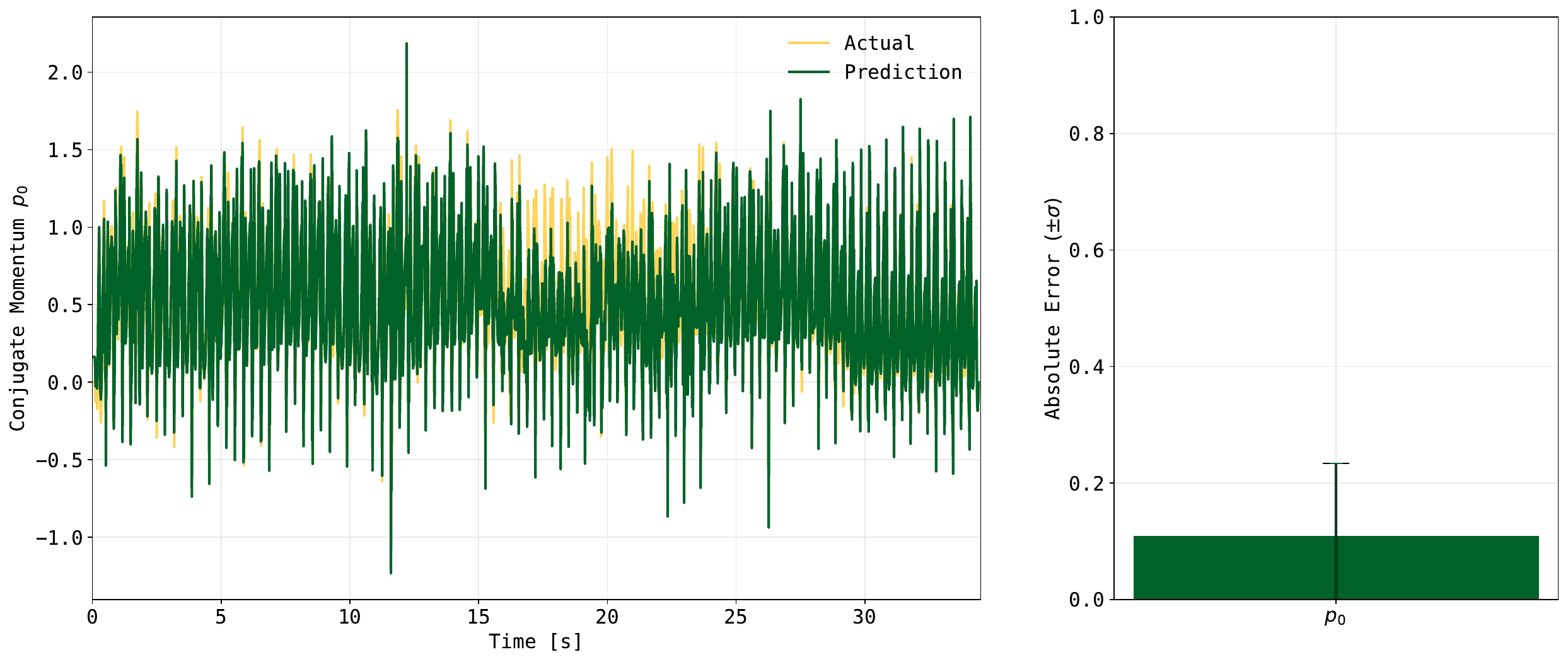}
    \caption{(Left) Predicted conjugate momentum (green) against the actual conjugate momentum (yellow). (Right) Absolute error for the conjugate momentum. }
    \label{fig:conjugate}
\end{figure*}

\subsection{Dirac Structures: The unifying language}

While the construction above is often introduced ad hoc, it can be stated elegantly in the language of \textit{Dirac structures}. A Dirac structure on a vector space $V$ is a maximal isotropic subspace $D\subset V\oplus V^{*}$ with respect to the symmetric pairing
$\langle(u,\alpha),(v,\beta)\rangle = \alpha(v) + \beta(u)$ \cite{courant1990dirac}. The graphs of closed 2-forms ($\operatorname{graph}\omega^{\flat}$) recover presymplectic geometry, whereas the graphs of bivectors ($\operatorname{graph}\pi^{\sharp}$) recover Poisson geometry; Dirac structures therefore {unify presymplectic and Poisson mechanics} within a single framework \cite{yoshimura2006dirac}. For constrained mechanical systems, the bundle:
~
\begin{align*}
D(\mathbf q) = \bigl\{ (v,\alpha) \in T_qQ \oplus T_q^{*}Q \mid \\
\alpha(v') = \omega(v,v') \;\text{for all}\; v' \in T_qQ \bigr\},
\end{align*}
captures simultaneously the admissible motions $v$ and the consistent constraint forces $\alpha$. Symplectification corresponds to lifting $D$ to the canonical graph, $\operatorname{graph}\widetilde\Omega^{\flat}$, on the extended space, after which every classical symplectic tool, including variational integrators and backward-error analysis, becomes available.

\subsection{Our Contributions}

Building on this geometric foundation, we make the following contributions:

\begin{enumerate}
\item \textbf{Presymplectification Network (PSN).}
We propose the first neural architecture that \emph{learns} the full presymplectification lift $\Psi_{\theta}:(\mathbf q,\mathbf p)\mapsto(\mathbf Q, \mathbf P)$, where the network outputs the Lagrange multipliers, their conjugate momenta, and the clock coordinate.
\item \textbf{Flow matching training.} The core module of the PSN consists of a model that combines a Gated-Recurrent Unit (GRU) network with a normalizing-flow-style velocity head, wrapped inside a differentiable implicit midpoint integration layer. This yields a map compatible with the learned Dirac lift \cite{cho2014learningphraserepresentationsusing, lipman2023flowmatchinggenerativemodeling}.
\item \textbf{Dynamics prediction module.} As a downstream task, we use SympNets to predict the dynamics of a complex physical system.
\end{enumerate}

In order to demonstrate, the effectiveness of this pipeline we choose to predict the locomotion dynamics of the ANYmal quadruped robot \cite{anymal}, a non-holonomic, multibody system with contact constraints that is often considered to be amongst the most challenging problems in robotics.

\section{Methods}

This section formalizes the learning task illustrated in Fig.~\ref{fig:framework_overview}, details the \textit{Presymplectification Network} (PSN) architecture, introduces our \textit{flow-matching} training objective, and describes the downstream \textit{SympNet} step predictor that together form the complete pipeline. Throughout, we write $\mathbf x=(\mathbf q,\mathbf p)\in T^{*}Q$ for the physical state and $\mathbf z=(\mathbf Q, \mathbf P)\in T^{*}\widetilde{Q}$ for its lifted counterpart.

\begin{figure*}[htb!]
    \includegraphics[width=1.0\textwidth]{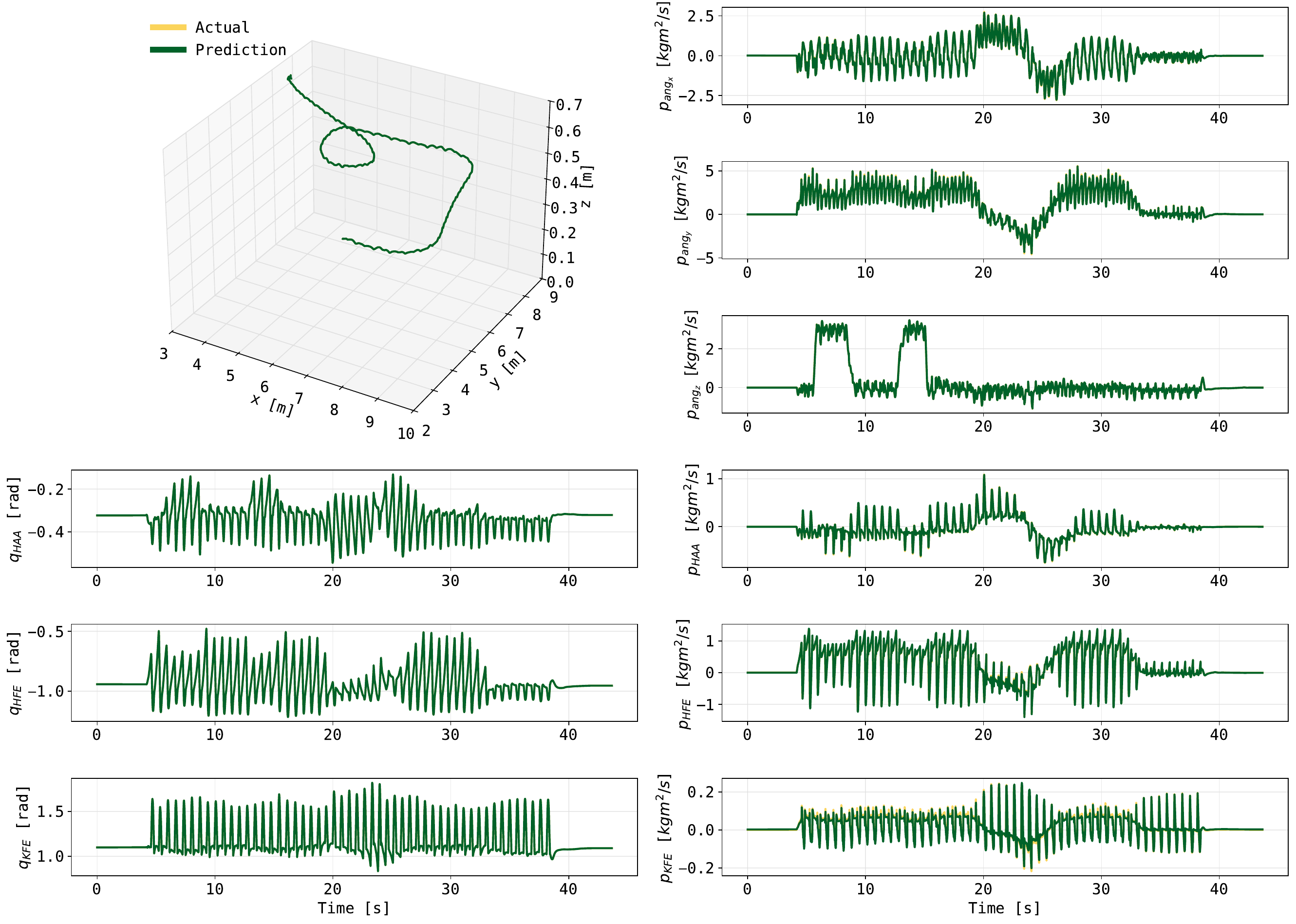}
    \caption{Inference Results of Dirac-Lifted dynamics prediction for the ANYmal quadruped. The 3D plot showcases the path of the quadruped's base-origin in space, with the corresponding angular momentum coordinates $p_{ang_{x}},\;p_{ang_{y}}$ and $p_{ang_{z}}$ on the right. The remainder lower-positioned plots illustrate the joint-angles and corresponding joint momenta for the Right Fore leg of the quadruped. The high overlap between the predictions (green) and the actual response (yellow), indicates the high accuracy and excellent performance of our Dirac-lifted dynamics prediction framework.}
    \label{fig:pred}
\end{figure*}

\subsection{Problem Statement}

The ANYmal \cite{anymal} is a torque-controlled quadrupedal robot engineered for agile locomotion and resilient interaction with unstructured environments. Unlike many legged platforms, it features relatively heavy limbs, making their inertial and Coriolis effects dynamically significant and thus, posing a substantial challenge for accurate modeling and control. However, our modular pipeline alleviates these difficulties by re-parametrizing the quadruped's dynamics in the lifted Hamiltonian phase-space ($T^{*}\widetilde{Q}$) and predict its dynamics on this lifted manifold. More specifically, the dynamics of the quadruped can be expressed by:
\begin{equation}
\begin{aligned}
    &\mathbf M(\mathbf q) \ddot{\mathbf q} + \mathbf C(\mathbf q, \dot{\mathbf q}) = \mathbf B\mathbf u + \mathbf J_c(\mathbf q)^\top \mathbf F_c,\\
    &s.t.: \phi_i(\mathbf q) = 0, \forall i \in \mathcal I_{c},
\end{aligned}
\end{equation}
where $\mathbf M(\mathbf q)$ represents the mass matrix, $\mathbf C(\mathbf q, \dot{\mathbf q})$ the nonlinear dynamics terms, $\mathbf B$ the control-input ($\mathbf u)$ selection matrix, $\mathbf J_c(\mathbf q)$ the concatenated contact Jacobians and $\mathbf F_c$ the contact wrenches. The scleronomic contact constraint $\phi_i(\mathbf q) = 0$ is active only for the feet ($i$) that are in contact, within the contact set $\mathcal I_c$. The configuration space $(Q)$ comprises a composite lie group $\mathbb{SE}(3)\times \mathbb R^{12}$, while its tangent space $TQ$ lives in the euclidean $\mathbb R^{18}$ space. After the Dirac lift, the system is characterized by a lifted 86-dimensional augmented phase-space with the base orientation represented as quaternion.

Given a sequence of timestamped states, $\mathcal{D}=\{(\mathbf q_t,\dot{\mathbf q}_t)\}_{t=0}^{T}$, collected from the simulation of a physical system (ANYmal in our case), our goal is twofold:

\begin{itemize}
\item \textbf{Lift learning:} learn a map $\Psi_\theta:T^{*}Q\to T^{*}\widetilde{Q}$ such that:
\begin{equation}
    (D\Psi_\theta)^{\top}\widetilde{\Omega}\,D\Psi_\theta=\Omega,
    \quad
    \Psi_\theta(\mathbf x)=(t,\mathbf q,\boldsymbol{\lambda},\;p_0,\mathbf p,\boldsymbol{\pi}),
\end{equation}
i.e.\ the pull-back of the canonical form is the original (possibly degenerate) form.
\item \textbf{Dynamics prediction:} learn a discrete flow, 
$\Phi_\phi:T^{*}\widetilde{Q}\to T^{*}\widetilde{Q}$,
that advances the lifted state over one step $\Delta$ while remaining symplectic, 
$\Phi_\phi^{*}\widetilde{\Omega}=\widetilde{\Omega}$.
\end{itemize}

It has to be noted that for multibody systems, such as quadrupeds, the lifted phase-space consists of interpretable physical quantities: the conjugate momentum $\mathbf p_0$ consists of the non-conservative energy parts of the system, hence dissipation and control-input energy, while the constraints $\boldsymbol{\pi}$ correspond to each contact constraint $\boldsymbol \phi(\mathbf q)$ and its Lagrange multipler $\boldsymbol{\lambda}$ to the contact wrenches $\mathbf F_c$. Finally the generalized momenta ($\mathbf p$) of the system can be calculated as $\mathbf M(\mathbf q) \dot{\mathbf q}$.

These intuitive connections are met in many dynamical systems, providing a tangible manifestation of the abstract Dirac procedure, which lifts the dissipative system to a higher dimensional space that becomes fully conservative.

\subsection{Presymplectification Network Architecture}\label{sec:arch}
\subsubsection{Encoder}
The encoder $\Psi_\theta$, given the original phase space $\mathbf x_t$, the control $\mathbf u$ and the re-parametrized time $t$, learns the flow to the target conjugate momentum $p_0$ using an \textit{inpainting-based} training technique. Given a lifted state $\hat{\mathbf z}_k = (t_k, p_{ctrl_k}, \mathbf x_k)$ Using a three-layer gated recurrent unit (GRU) followed by linear heads:
\begin{align}
\mathbf h_k &= \mathrm{GRU}_\theta(\hat{\mathbf z}_k, \mathbf h_{k-1})\; \forall\;k\in [0, t-1], \\
\mathbf v_t &= \mathbf W_1 \hat{\mathbf z}_t + \mathbf W_2 \mathbf h_{t-1} + \mathbf b,
\end{align}
where $\mathbf v_k$ corresponds to the velocity flow.

\subsubsection{Implicit Midpoint Layer}
We use a continuous-time GRU formulation by employing an
\textit{implicit midpoint layer}. Given the lifted state $\hat{\mathbf z}_t$
we solve:
\begin{equation}
\hat{\mathbf z}_{t+\Delta} =  \hat{\mathbf z}_{t} + \mathbf v_{t+\Delta/2} \Delta
\label{eq:implicit},
\end{equation}
where the velocities $\mathbf v$ are predicted using the encoder $E_{\theta}$.

\subsection{Flow-Matching Objective}\label{sec:flowmatching}
Classical sequence-to-sequence losses require ground-truth
$\widetilde{H}$ or contact forces, which are hard to measure.
Instead we minimize the discrepancy between
the \emph{data velocity field} $\mathbf v^*$ and the velocity that PSN
induces after projection back to $T^{*}Q$. Hence, the flow-matching loss is:
\begin{equation}
\mathcal{L}_{\text{FM}}
=
\frac{1}{|\mathcal{D}|}
\sum_{t}
\norm{
\mathbf v^*
- \Pi(\mathbf v_t)
}_2^{2},
\label{eq:flowmatching}
\end{equation}
where $\Pi$ is a projection operator that maps the full augmented velocity $\mathbf v_t$ to the physical phase space coordinates,
which reduces to, and generalizes, the
exact transport equation used by \cite{lipman2023flowmatchinggenerativemodeling}.

\subsection{SympNet Step Predictor}
Once the PSN has been trained,
we attach a lightweight SympNet \cite{jin2020sympnets} that
maps the \emph{lifted} coordinates forward one discrete step. Note that in this lifted space ($\mathbf z$), the constraints $\boldsymbol \pi$ and the corresponding multiplers $\boldsymbol \lambda$ are concatenated to the flow-matching's phase-space $\hat{\mathbf z}$:
\begin{equation}
\mathbf z_{t+\Delta} = S_\phi\bigl(\mathbf z_t\bigr),
\qquad
S_\phi^{*}\widetilde{\Omega}=\widetilde{\Omega}.
\label{eq:sympnet}
\end{equation}
Equation~\eqref{eq:sympnet} inherits symplecticity by design as SympNet is constructed as a composition of coupled $S$-blocks which calculate the velocity in tangent space, while mapping each with an exact Lie-Trotter symplectic map. The SympNet is trained with a standard one-step prediction loss:
\begin{equation}
\mathcal{L}_{\text{pred}}
=
\frac{1}{|\mathcal{D}|}
\sum_{t}
\norm{\Pi(\mathbf z_{t+\Delta})
\ominus \mathbf x_{t+\Delta}
}_2^{2},
\label{eq:pred}
\end{equation}
while the PSN parameters are \emph{frozen},
guaranteeing that the learned lift remains unchanged. The $\ominus$ operator computes the difference between the projected predicted state $\Pi(\mathbf z_{t+\Delta})$ and the ground-truth physical state $\mathbf x_{t+\Delta}$, appropriately handling the composite Lie-Group of the floating-base quadruped.

\section{Results}
In Fig.~\ref{fig:conjugate}, we benchmark the performance of PSN in predicting the conjugate momentum $p_0$ against a ground truth dataset from out-of-distribution initial conditions for the quadruped ANYmal during simulation. The GRU constitutes a good trade-off between scalability and inference complexity, while for more complex systems, transformer-based architectures~\cite{vaswani2017attention} may be more suitable.

The results of the integration downstream task are showcased in Fig.~\ref{fig:pred}. The actual (yellow) and prediction (green) lines overlap almost perfectly, proving the effectiveness and accuracy of our Dirac-lifted framework for dynamics prediction of complex systems. The plots showcase the highly nonlinear locomotion dynamics of the robot base and of one of the robot's legs.

\section{Future Work and Conclusion}
Despite the strong empirical performance of the pipeline, several research avenues remain open:

\begin{itemize}
\item \textbf{Fully symplectic flow matching.}
Our present objective (Eq.~\ref{eq:flowmatching}) matches the \emph{projected} velocity field on the physical phase space. An immediate extension is to lift the flow‐matching loss itself, enforcing it directly on $T^* \widetilde{Q}$ while preserving symplecticity at each step. Such \emph{symplectic flow matching} would dispense with the implicit midpoint layer and allow the development of score-based generative models on symplectic manifolds.
\item \textbf{Multi-step prediction.}
Moving beyond one-step rollouts, we plan to train sequence models, e.g.\ symplectic Transformers or recurrent SympNets, on the lifted coordinates so that the network can forecast an entire horizon in a forward pass as a downstream task.
\item \textbf{Scalability to articulated collectives.}
Finally, evaluating our pipeline on modular manipulators and multi-robot swarms, where the constraint topology varies over time, will stress-test memory efficiency and training stability.
\end{itemize}
We have introduced {Presymplectification Networks}, the first deep-learning framework that learns Dirac’s presymplectification lift, and coupled it with symplectic prediction to model contact-rich robotics. By lifting data to a non-degenerate symplectic manifold, enforcing flow-matching consistency, and forecasting with a lightweight SympNet, our pipeline preserves energy, momentum, and holonomic constraints while performing well on a challenging quadruped robot dataset. These results demonstrate that presymplectification can be learned end-to-end and exploited for practical robotics. Our future efforts aim at symplectic flow matching, multi-step prediction, and automated contact discovery,  with the aim of bringing us even closer to data-driven models that match the rigor and reliability of first-principles mechanics.


\section*{Acknowledgements}
The authors would like to thank Vassil Atanassov for the useful discussions. A.P is supported by University of Oxford’s Clarendon Fund. P.V. is supported by the United States Army Research Office under Award No. W911NF-21-S-0009-2. N.A. acknowledges support from the European Research Council (grant agreement 948932) and the Royal Society (URF-R1-191150).

\bibliographystyle{ieeetr}
\bibliography{references}

\end{document}